# Segmentation of ultrasound images of thyroid nodule for assisting fine needle aspiration cytology

Jie Zhao[1*]
[*] Corresponding author
Email: Zhaojie_hbu@126.com

Wei Zheng[1]
Email: weizheng799@yahoo.com

Li Zhang[1]
Email: zhaojie_hbu@126.com

Hua Tian[1]
Email: zhaojie_hbu@126.com

[1] College of Electronic and Information Engineering of Hebei University, Baoding 071002, China

## Abstract

The incidence of thyroid nodule is very high and generally increases with the age. Thyroid nodule may presage the emergence of thyroid cancer. Most thyroid nodules are asymptomatic which makes thyroid cancer different from other cancers. The thyroid nodule can be completely cured if detected early. Therefore, it is necessary to correctly classify the thyroid nodule to be benign or malignant. Fine needle aspiration cytology is a recognized early diagnosis method of thyroid nodule. There are still some limitations in the fine needle aspiration cytology, such as the difficulty in location and the insufficient cytology specimen. The accuracy of ultrasound diagnosis of thyroid nodule improves constantly, and it has become the first choice for auxiliary examination of thyroid nodular disease. If we could combine medical imaging technology and fine needle aspiration cytology, the diagnostic rate of thyroid nodule would be improved significantly.

The properties of ultrasound, such as echo, shadow, and reflection, will degrade the image quality, which makes it difficult to recognize the edges for physicians. Image segmentation technique based on graph theory has become a research hotspot at present. Normalized cut (Ncut) is a representative one, whose biggest advantage is not prone to small region segmentation but suitable for segmentation of feature parts of medical image. However, how to solve the normalized cut has become a problem, which needs large memory capacity and heavy calculation of weight matrix. It always generates over segmentation or less segmentation which leads to inaccurate in the segmentation.

The speckle noise produced in the formation process of B ultrasound image of thyroid tumor makes the quality of the image deteriorate. In the light of this characteristic, we combine the anisotropic diffusion model with the normalized cut in this paper. After the enhancement of anisotropic diffusion model, it removes the noise in the B ultrasound image while preserves the important edges and local details. This reduces the amount of computation in constructing the weight matrix of the improved normalized cut and improves the accuracy of the final segmentation results. The feasibility of the method is proved by the experimental results.

## Introduction

Epidemiological studies show that the incidence of thyroid nodule is very high and increases with the age. It may presage the emergence of thyroid cancer [1]. Thyroid carcinoma is different from other thyroid cancer. It can be completely cured if detected early. Therefore, it is very necessary to correctly classify the thyroid nodule [2] to be benign or malignant. Fine needle aspiration cytology (FNAC) [3] is a recognized early diagnostic method of thyroid nodule with high success rate, but it still has some limitations. Therefore, even if the patients whose results is invisible through FNAC, they also can't completely be ruled out the possibility of a tumor.

Thyroid ultrasound technology [4] can ensure that you get a lot of information about thyroid nodule before the operation. Through the ultrasound images, we can locate the position of the thyroid nodule, measure the size, and decide whether an operation is needed or not. In ultrasound images, in addition to the echo characteristics, nodules in degree, there are some other ultrasound characteristics which can also be acted as a judgment indicator which shows possibility of nodular malignant, such as the shape and contour of nodules. The exact boundary detection of ultrasound images [5] will provide accuracy position for pierce, but it exists a granular pattern which called spots in the ultrasound images because of the impact of imaging principle. In addition, the properties of echo, shadow, and reflection of ultrasonic will degrade the image quality. This image quality degradation caused by the nature of ultrasonic image makes it difficult to recognize its edges accurately even for an experienced physician. Especially it is very difficult to complete the nodules and tracheal of nodules positioning area of the regional segmentation.

At present, the segmentation method which is widely used in the clinical application of ultrasound imaging systems is based on the threshold value method or the doctor manual segmentation method. Although the implement method of threshold segmentation is convenient and simple, inevitably, the speckle noise and texture in the ultrasound image make the method difficult to obtain satisfactory results [6]. In the above segmentation method, manual segmentation method is relatively easy to implement and the result is also easy to accept, but the heavy workload and long time tend to make doctors and patients difficult to accept. Using computer to implement automatic, semi-automatic segmentation method is the ideal choice for ultrasound image segmentation in clinical application. As an important branch of image segmentation, the segmentation of medical ultrasound image almost covers all existing segmentation techniques. Aarnink et al. use the nonlinear Laplace filter to implement segmentation of prostate ultrasound images automatically [7]. Fan et al. use the nonlinear wavelet threshold method to detect the boundary which is formed by lumen-intima - the inner wall and outer membrane of implantable ultrasound images. Yoshida et al. do a more in-depth study of medical ultrasound image segmentation which is based on the active contour models. Lee et al. take advantage of the dynamic programming algorithm to segment different medical ultrasound image and have achieved good segmentation results [8]. Yan Jiayong et al. use Active Contour Model which is based on Gradient Vector Flow Law has achieved a certain effect on soft tissue tumors segmentation of ultrasound image. Cvancarova et al. put forward Snake model of ultrasonic image segmentation method based on GVF algorithm [9]. According to characteristics of ultrasound cardiac image with noise, fuzzy boundaries, uneven distribution of grayscale, Zhang Et al. propose polarity filtering and edge sharpening, and then use CV Snake model to segment, extracted ventricular boundary finally. Yan Et al. use the average edge energy of zero level set curves to control energy evolution speed and segmentation results by studying Chan-Vese level set method. Liu Jinzhu et al. put forward ultrasound image classification method of fatty liver which is based on threshold segmentation to analysis of the characteristics of the lesions and non-lesion tissue in ultrasound images in detail. In the end, they point out the main problem in the ultrasound image segmentation.Yu Jiali et al. propose medical ultrasound image segmentation based on a random walk. Through solving sparse, symmetric and positive definite equations of linear system to obtain the solution of the problem, and then realize the segmentation of medical ultrasound image.

Since ultrasonic image exists serious artifact and noise, at the same time the target has weak boundary or boundary breakpoints, which makes method based on boundary difficult to segment correctly. The segmentation algorithm based on texture needs a predefined image mode, so the effect is not good in the ultrasonic image segmentation, and often need to combine with other knowledge to improve the segmentation results. The method based on the model can effectively segment target ,use the homogeneous area statistical information to structure energy function, and search minimization optimal solution in the global range . But the image segmentation effect is not ideal for the target which is disorderly or uneven distribution of grayscale. Therefore, the present algorithm exists certain flaws in dealing with speckle noise and weak boundary, and most of the algorithms need to manually draw the outline of the initial contour which is close to the real boundary of the target, so it is difficult to get wide application in clinical. According to the above problems which exist in the ultrasonic image segmentation in this paper and combined with the characteristics that the thyroid ultrasonic image is seriously polluted by the speckle noise and images is usually fuzzy, we propose an improved image segmentation algorithm based on normalized cut is, combining homomorphism filtering, the anisotropic diffusion model, fractional differential into normalized cut. The speckle noise is removed, important edge details are preserved, and the amount of computation of weight matrix is reduced. And the algorithm is compared with some traditional segmentation method including edge detecting, threshold segmentation, region splitting and merging and some modern segmentation method including watershed segmentation, active contour model, and graph segmentation method [10]. Simulation experiment show that only the improved segmentation method based on normalized cut can segment the important parts, such as tumor, thyroid, and windpipe and so on. The segmented parts is the important reference value for the doctors to diagnose thyroid tumor in clinical.

The remainder of this paper is organized as follows. In Section 2, we analyze the characteristics of thyroid nodule ultrasound image .The traditional method for segmentation of thyroid ultrasound image is introduced in Section 3. Ultrasound image segmentation based on Ncut is introduced in Section 4. The proposed method and the implementation and results are given in

Section 5. This paper is summarized in Section 6.

## Characteristics analysis of thyroid nodule ultrasound image

The following group of thyroid ultrasound images shown in Figure 1 is provided by Affiliated Hospital of Hebei University.

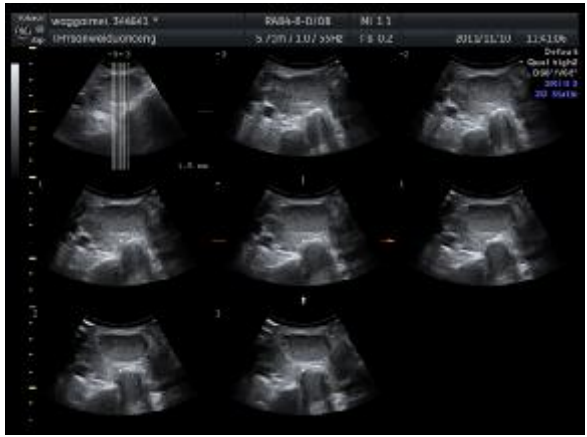

**Figure 1 The original ultrasound image of thyroid of one patient**

We show a group of three-dimensional ultrasound images of the SPECT images in Figure 1. The width of the image is 640 pixels, height 480 pixels, with 150 DPI resolutions both in horizon and vertical, and a depth of 24 bits. Such groups of image must be preprocessed before image segmentation, which requires us to transform SPECT image into independent images before processing.

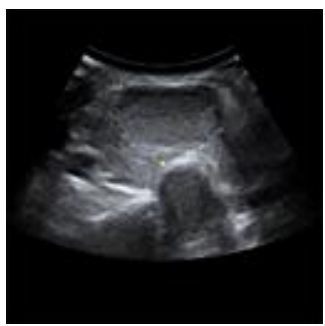

**Figure 2 The single layer of thyroid ultrasound image after cutting**

The size of the cut image is 128 by 128, 150 DPI. The histogram is an important statistical characteristic of the image. It represents the statistical relationship between the probability of each gray level and the gray-scale, provides the whole distribution of the grey value. The histogram of original thyroid ultrasound image of Figure 2 has shown in Figure 3.

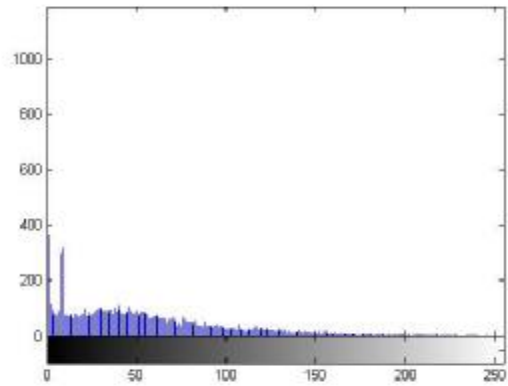

**Figure 3 The histogram of thyroid ultrasound image**

As we can see from the grey distribution of Figure 3, the grey composition of the image is mainly concentrated in dark side, basically in the gray level of 180 or so, and do not cover the gray level to get all the range, so the dynamic range is very small.

## The traditional method for segmentation of thyroid ultrasound image

### *The method of edge detection*

Edge detection is the most basic image segmentation method. In generally, between the different regions appears necessarily in different gray mutations namely edge. We can detect the first derivative maximum or second order derivative zero to test the edge. In general, different edge detection operator [11] templates were designed and used to complete the image convolution. Commonly, the first derivative operators include gradient operator, Prewitt operator, and Sobel operator. The second order derivative operators, such as Laplace operator, are particularly sensitive to edge information and noise, so the unnecessary noise should be removed before the edge detection.

In order to reduce the influence of noise to the image, we usually conduct image filtering before the derivation. Commonly we use Canny operator or LOG operator. LOG operator is developed on the basis of the Laplace operator. It firstly uses a Gaussian function to smooth the image, and then use the Laplace operator to detect the edge of the smoothed image. Similarly, Canny operator first uses Gaussian filter to smooth the image, and then adopts the maximum inhibition and double threshold to detect the edge of the gradient image, which can lead to good detection results and high precision.

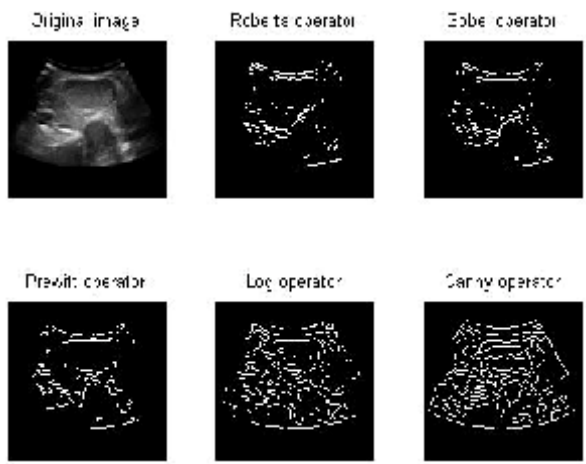

**Figure 4 The edge detection of thyroid ultrasound image**

From the results of traditional edge detection we can see that the edge is discontinuous, and a lot of false edges were detected.

### *The method of threshold value*

Using the difference of gray image level between the target object and the background area, we divide gray image into target area and background area, then use one or several threshold value of gray image to divide gray image into several parts. This method is called image threshold segmentation, which is widely used in image segmentation. That is to say, the threshold segmentation method is based on the assumption that similar pixels have similar gray values, not similar pixels have great differences between gray value, which is reflected on the histogram that the different classes correspond to different peaks. We always select the valley between the two peaks as the segmentation threshold during the image segmentation, which will separate each peak, and then complete the image segmentation. It is thus clear that the key to threshold segmentation is to find the optimal threshold so that we can separate two types of target. However, not all images have the obvious bimodal or multi-peak in histogram, so the choice of thresholds is becoming more and more difficult. In this case, many other improved methods to determine the threshold [12] have been put forward, such as the approach based on transition zone, the changed threshold method of pixel spatial location information, the threshold method combined with the connectivity information and so on.

The threshold segmentation is simple for the different types of objects that have big difference on gray value or other characteristic values, which would be very effective in image segmentation.

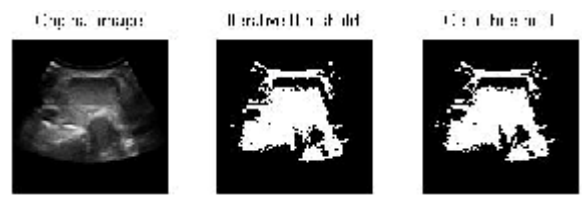

**Figure 5 Threshold segmentation of thyroid ultrasonic images**

From the results, we can see that the histogram of original thyroid ultrasound image is not bimodal image, so the result is wrong. In conclusion, threshold segmentation method can not be adopted in this situation.

## *Region splitting and merging method*

Region growing method mainly considers the relationship between the pixels and it's spatially neighborhood pixels. It is a way of extracting connected region in a image according to predefined standard. Specific approach is as follows: first, identify one or more seed points as the starting point(s) of growth, and then merge the pixels which have the same or similar characteristics with the seed pixels in the neighborhood into the area of the seed pixels. Regard these new pixels as new seed pixels to continue the process above until there is no pixels which suffice the conditions. The core of the region growing method is the selection of seed point and the measure of regional similarity [13].

Region growing method has the advantage of simple calculation, and it also considers the pixel similarity and spatial neighborhood, thus it can effectively eliminate isolated noise points, and it is especially suitable for the segmentation of small structures, such as tumor and scar detection and segmentation. The disadvantage is that we have to manually implant a seed point to every area we need to extract, and it is particularly sensitive to the selection of seed points, for different seed points may get very different segmentation results. Meanwhile, this method is particularly sensitive to noise and it is easy to cause inanity in the area. Regional split and merge method is developed on the basis of region growth method, it does not need to artificially determine the seed point and has overcome the defects that the region growing method needs to manually select the seed points. It splits and merges the whole image at the same time according to some consistency criterion, not liking the region growing method to start from a single pixel; splitting and merging focus on the design of the split and merge guidelines. The split and merge algorithm is effective in segmentation of complex scene images, but there are also shortcomings. It generally cannot reach pixel-level segmentation accuracy, because the pixel-level split and merge would increase the algorithm's time complexity, and also easily form the meaningless area.

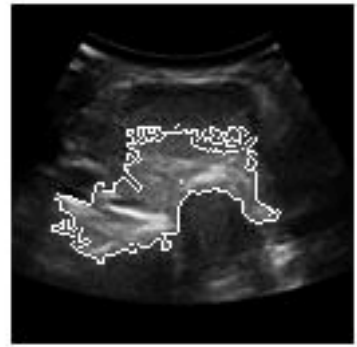

**Figure 6 The split and merge result of thyroid ultrasound images**

From the split and merge results of the ultrasound images, we can see that it splits neither the nodules nor the trachea, so the result has no practical significance.

## *Watershed segmentation method*

Morphological watershed algorithm is an algorithm which is based on region segmentation. The basic idea is to simulate the process that water flow submerges landform, to split the different areas through forming the dams between different regions. Watershed transform regards gray scale image as a geomorphic surface, and assumes to make a hole in the surface of each minimum area, water will slowly immerse in the surface from these holes, and starting from the minimum of lowest point, water will gradually submerge the catchment basin of the image. In addition, at a certain point, when the water from two different minimums increasingly rises to come together, it will build a dam at this point, at the end of the soaking process; each region minimum is surrounded by the dam of corresponding catchment basin, all the dam collection constitutes the watershed, which divide the image we input into different regions.

The whole watershed process can be described by mathematics:

Let $M_1, M_2,\ldots, M_R$ represent a minimal area of the image $f(x,y)$, $C(M_i)$ represents the catchment basin related to the minimal area $M_i$, min and max represent Gray-scale maximum and minimum of the image $f(x,y)$ respectively. Suppose that $T[n]$ represents a set in which all points $(s,t)$ suffice $g(s,t)<n$, that is to say: $T[n] = \{(s, t)|g(s, t) < n\}$. From a geometric perspective, $T[n]$ is the set of points located below plane $g(s,t)=n$ in image $f(x,y)$, that is to say, $n$ represents the immersion depth of step $n$. For a given catchment basin, in the step $n$, it will appear a certain degree of immersion (may not appear). Suppose that in step $n$, the minimal area $M_i$ is immersed, let $C_n(M_i)$ represent a part of the catchment related to minimal area $M_i$, which is the horizontal surface area formed in the catchment basin $C_n(M_i)$, when the immersion depth is $n$. In order to facilitate the discussion, we may regard $C_n(M_i)$ as a two value image, which can be represented by the following equation: $C_n(M_i) = C(M_i) \cap T[n]$. In other words, if it is at the position $(x,y)$, suffice $(x, y) \in C(M_i)$ and $(x, y) \in T[n]$, than $C_n(M_i) = 1$, otherwise $C_n(M_i) = 0$. If the Gray value of the minimal area $M_i$ is $n$, than in the step $n+1$, the immersed part of the catchment and the minimal area are exactly the same, that is $C_{n+1}(M_i) = M_i$. Suppose that $C[n]$ represents the union of the immersed part of all the catchment basin, that is

$$C[n] = \bigcup_{i=1}^{R} C_n(M_i)$$

, than $C[\max + 1]$ is the union of all the catchment, that is to say:

$$C[\max+1] = \bigcup_{i=1}^{R} C_n(M_i).$$

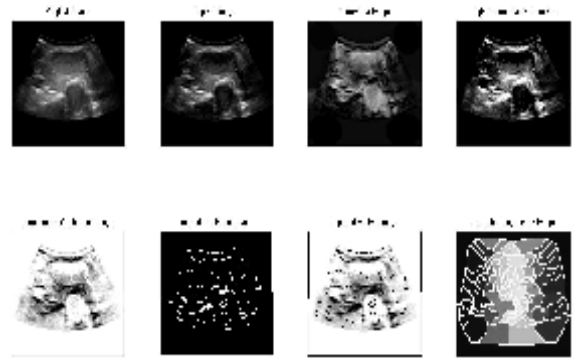

**Figure 7 Watershed segmentation of thyroid ultrasound images**

From the final results of the watershed segmentation, we can see that the phenomenon of over-segmentation is quite serious. It does not accurately segment nodules or trachea.

## Active contour model method

Curve evolution model segmentation method unifies the image, the initial contour, the target contour and the constraint condition and sets the initial curves and curved surface in image space and defines the internal energy related to curve or curved surface shape and the external energy related to the image. The internal energy controls the smooth and continuity of curve or curved surface, and the external energy relates to edge characteristics. In the interaction of internal energy and external energy, the contour deforms, so we can get the continuous edge images finally. This model, when segmenting an image, makes full use of the prior knowledge about the position, size, and shape of the interested region and the inherent information of medical image to reflect this prior knowledge in the energy functional form. It links with image data in a dynamic way, the energy function acts as a measure about coincidence degree between priori model and image data, minimization of energy function makes the final result of the curve evolution that contour curve approaches target contour. In addition, active contour model provides an interactive operating mechanism, which brings the professional knowledge into image analysis to significantly improve the robustness of the algorithm. Active contour model has now been widely used in object recognition, computer vision and other fields.

At present, according to the basic expression method of the curve, active contour model is divided into parameter-based active contour line model and geometry-based active contour line mode. Parameter active contour line model is also known as the Snake model. The basic idea of the Snakes model for image segmentation is to gain the edge of the image through deforming the initial curve. The basic process is firstly to delineate the detected target block in the image plane, then evolve and deform the closed curve, so that it can automatic stop when it arrives at the target boundary. The deformation process is obtained through minimizing an energy function. It is mainly composed of two parts, one part controls the smoothness of curve, another part forces curves to tend to the edge of the image. But the model has three disadvantages: first, it is sensitive to the initial curve location; second, curve in the course of evolution easily falls into local minimum point because of the non-convexity of the energy functional, making the segmentation failure; third, the topology of curves does not change in the course of evolution. Therefore, in the original model, we must pre-define an initial curve which surrounds it for each target object in the image, so that we can get the correct segmentation results. But this is a cumbersome and time-consuming work.

Based on the above problems, people present the geometric active contour line model to overcome the shortcomings of the parameter active contour line model. Geometric active contour line model concept was first proposed by Caselles in 1993. It is better to overcome the defects that the parametric model cannot handle topological changes. The biggest difference with the parametric active contour model is that it introduces level set method. Its initial contour moves toward the target edge under the impetus of contour curve geometric characteristics, and it has nothing to do with the parameters characteristic of the contour. It avoids the deficiency that parametric active contour model must repeat parametric curve and can automatically handle the question about curve topology changes.

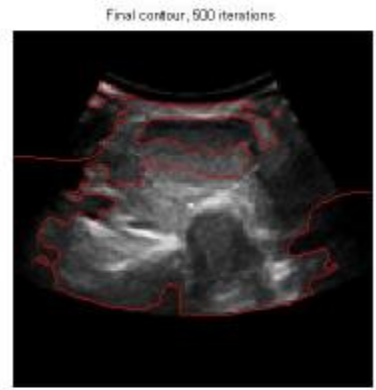

**Figure 8 The geometric active contour model segmentation of thyroid ultrasound image**

From the segmentation results of the geometric contour model method, we can see that even after 500 iterations, it does not get a meaningful segmentation area due to the low contrast of the ultrasound image.

## A segmentation method based on graph theory

Among many image segmentation algorithms, the Graph-based segmentation method shows a powerful advantage, which can effectively combine the image gray level, texture, color and other information of the image to achieve a satisfactory segmentation results. In addition, many mature and perfect classical algorithms of Graph-based segmentation methods provide a powerful computational tool. So the algorithms attract widely attentions in recent years.

The algorithm based on the cut value of image [14] is one of the most important kinds of the Graph-based segmentation method. The theory of this method regards the image as an undirected weights graph, the vertex of the image corresponds to the pixels or regions of the image, the weights of edge reflects the similarity between two pixels or regions. We solve the extreme value of the objective function which is defined to use a certain cut value to achieve segmentation. There are several classic algorithms based on the cut value of image. The Min Cut is presented earlier as a optimization criterion, it can obtain better segmentation result for some images. However, it will segment isolated points or small regions. Because the base value Cut(A, A ) will increase along with the increase of the edge numbers between A and A. Then many optimization criterions are proposed to solve this disadvantage. They average the cut value from different angles, including averaging the volume of A ( isoperimetric segmentation, normalized cut value ) and the potential of A (average the cut value ). Ncut is a kind of them, its biggest advantage is not prone to small region segmentation, and it suits to segment the of feature parts of medical image.

## Ultrasound image segmentation based on Ncut

In graph theory, a image is a method to describe the relationship between things. Image represents for $G = (V, E, W)$, $V$ is the set of all the nodes in the graph, E is the set of the edge with connecting the two nodes, $W$ is composed by the matrix $W_{ij}$, $W_{ij}$ represents similarity between the two nodes. Assume that divided graph $G$ into two disjoint subsets $A$ and $B$, then $A \cup B = V$, $A \cap B = \emptyset$, so the dissimilarity between two subsets can be expressed as the cut value (1):

$$Cut(A,B) = \sum_{\substack{i \in A, j \in B \\ (i,j) \in E}} w_{ij} \quad (1)$$

The best effect of image segmentation is to make the cut value minimum. Ncut [15] uses the normalized cut value as criterion:

$$Ncut(A,B) = Cut(A,B)(\frac{1}{Vol(A)} + \frac{1}{Vol(B)}) \quad (2)$$

$$Vol(A) = assoc(A,V) = \sum_{u \in A, v \in V} w(u,v) \quad (3)$$

$$Vol(B) = assoc(B,V) = \sum_{u \in B, v \in V} w(u,v) \quad (4)$$

The ultrasound images segmented by the normalized cut criterion are shown in Figure 9.

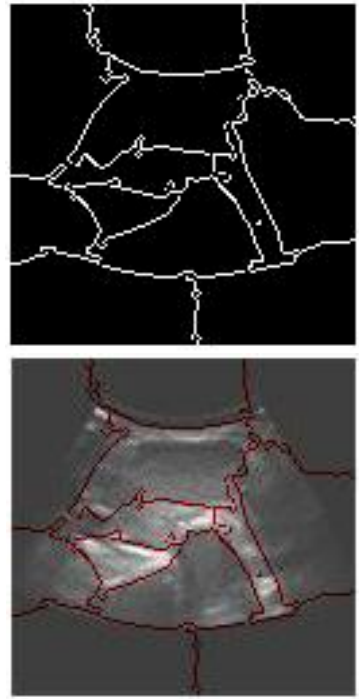

**Figure 9 The segmentation results of normalized cut**

From the results, we can see that the normalized cut produces less segmentation phenomenon, the divided areas have no practical significance, nodules and tracheal have not been segmented.

# The ultrasonic image segmentation based on improved normalized cut

## *The algorithmic principle*

Compare Figure 2 with Figure 3, the numbers of black pixels are large in grayscale distribution of the original image. To obtain an image which can segment and recognize the following image better, the dynamic range of the image must be compressed meanwhile the contrast of the image must be improved. Homomorphism filter is a method that the brightness range of image is compressed and the contrast of image is improved simultaneously in the frequency domain. The thyroid ultrasound images are processed by homomorphism filter showing as Figure 10.

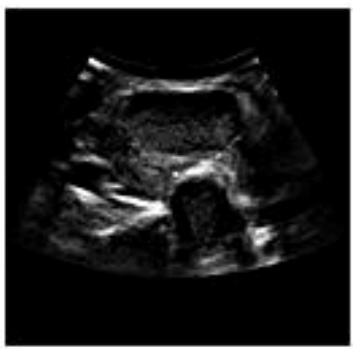

**Figure 10 Ultrasound image processed by the homomorphism filter**

The general image denoising is isotropic diffusion which removes noise while makes the boundary fuzzy simultaneously. The biggest characteristic of the anisotropic diffusion is that it is a selective smooth process which is not restricted in uniform regions but is limited in crossing boundary part. So noise and some irrelevant details are smoothly away, which can effectively achieve the smooth image edge. Gradient calculation based on the fractional differential can nonlinearity keep the low frequency components of signal and nonlinearity strengthens its high frequency texture detailed information [16]. So, this paper brings anisotropic diffusion filter model and fractional differential in normalized cut for processing image segmentation, as it's pre-treatment, it obtains the accurate segmentation results of thyroid ultrasound images.

## *Image enhancement based on anisotropic diffusion model*

Anisotropic diffusion model is structured by utilizing transformation of local coordinates, first second order derivative method of edge local details and hyperbolic tangent function combining anisotropic diffusion equations.

Anisotropic diffusion model derives from thermo diffusion equation (7), thermo diffusion equation is following as (6):

$$\vec{q} = -\vec{D} \cdot \nabla u \qquad (6)$$

The symbol $\vec{q}$ means heat flux fields, the symbol *u* means gradient field of temperature, and the symbol *D* means heat conductivity. From conservation of energy principle, thermal energy differential form is following as (7):

$$\frac{\partial u}{\partial t} = div(\vec{D} \cdot \nabla u) \qquad (7)$$

In image processing, the every bit value *u* of temperature field in planar region is regarded as grey value of this point of image, thermal diffusion process changes into the denoising processing of image. But the diffusion behavior of controlling every image bit should be done in local coordinate system, utilizing the coordinate transformation transform *X-Y* to local coordinate *M-N*.

Conversion expressing is following as (8),

$$\begin{pmatrix} m \\ n \end{pmatrix} = \frac{1}{|\nabla u|} \begin{pmatrix} u_x & u_y \\ -u_y & u_x \end{pmatrix} \begin{pmatrix} x \\ y \end{pmatrix} \quad (8)$$

The image $u$ gradient in local coordinate system is $[U_m, U_n]$, along $M, N$ direction, the diffusion coefficients are $f_1(x, y, t), f_2(x, y, t)$

$$q = -\begin{bmatrix} f_1 & 0 \\ 0 & f_2 \end{bmatrix} \cdot \begin{bmatrix} u_m \\ u_n \end{bmatrix} \quad (9)$$

The diffusion equation:

$$\frac{\partial u}{\partial t} = -div(q) = \frac{\partial(f_1 \cdot u_m)}{\partial m} + \frac{\partial(f_2 \cdot u_n)}{\partial n} \quad (10)$$

In any pixel of image, the diffusion coefficients are the variables of time. Diffusion equation is rewritten to:

$$\frac{\partial u}{\partial t} = div(D \cdot \nabla u) = f_1(x, y, t) u_{mm} + f_2(x, y, t) \quad (11)$$

Function (11) is the denoising anisotropic diffusion model, $f_1(x, y, t)$, $f_2(x, y, t)$ are smooth coefficients. In order to achieve the edge smooth, it must add edge enhancement in this model.

Because hyperbolic tangent function can gently control increasing and decreasing of the grey level of image edge on both sides of the center, reducing the edge width to strengthen edge. Therefore, the anisotropic diffusion model of finishing denoising and edge enhancement is (12),

$$\frac{\partial u}{\partial t} = a(x, y, t) div(D \cdot \nabla u) - b(x, y, t) f_3(x, y, t) th(l v_{mm}) |u_m| \quad (12)$$

$v = G_t \quad u; \quad th(l v_{mm}) = (e^{l v_{mm}} - e^{-l v_{mm}})/(e^{l v_{mm}} + e^{-l v_{mm}})$, $th(l v_{mm})$ are hyperbolic tangent function. The symbols $\alpha(x, y, t)$, $\beta(x, y, t)$ are anisotropic diffusion and edge enhancement respectively. The symbol $f_3(x, y, t)$ means edge enhance coefficient. The symbol $l$ means control curve slope. The symbol $G_t$ means Gaussian smooth function.

In view of above factors and human visual cover effect, the diffusion coefficient is:

$$f_1(x, y, t) = 1 / (1 + a|v_m|^2 + b|v_{mm}|^2) \quad (13)$$

$$f_2(x, y, t) = 1 / \sqrt{1 + a|v_m|^2 + b|v_{mm}|^2} \quad (14)$$

$$f_3 = 1 - 1/(1 + c|v_m|^2) \quad (15)$$

Coefficient $a$ controls the anisotropic diffusion for keeping edge and the local details, coefficient $b$ controls the anisotropic diffusion for keeping the article light, coefficient $c$ controls selectively the area of edge enhancement. The image enhanced by anisotropic diffusion is as Figure 11:

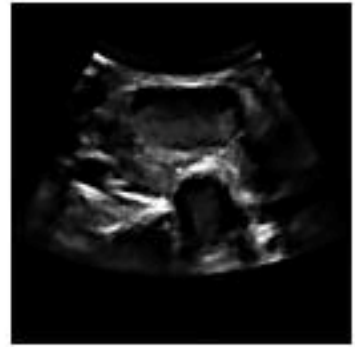

**Figure 11 The image after the anisotropic diffusion**

*The gradient image of ultrasound images based on Fractional Differential*

If the duration of one source $f(t)$ is $t \in [a,t]$, then parting the duration to equal part as unit interval $h=1$ so $n = [\frac{t-a}{h}]^{h=1} = [t-a]$, then derived difference expression of one source $f(t)$ fractional derivatives (16),

$$\frac{d^v f(t)}{dt^v} \approx f(t)+(-v)f(t-1)+\frac{(-v)(-v+1)}{2}f(t-2)+$$
$$\frac{(-v)(-v+1)}{6}f(t-2)+\frac{(-v)(-v+1)(-v+2)}{6}f(t-3)+ \quad (16)$$
$$\mathbf{L}+\frac{\Gamma(-v+1)}{n!\Gamma(-v+n+1)}f(t-n)$$

Generally speaking, in the image $f$ of $M \times N$, we use the filter masking of $m \times n$ linear filtering according to (17):

$$g(x,y) = \sum_{s=-a}^{a}\sum_{t=-b}^{b} w(s,t)f(x+s, y+t) \quad (17)$$

$a = (m-1)/2$, $b = (n-1)/2$, $x = 0, 1, 2, \ldots, M-1$, $y = 0, 1, 2, \ldots, N-1$. In allusion to the characteristics of thyroid cancer in the ultrasound images, The Mask operator of fractional differential is shown in Table 1.

**Table 1 Fractional differential operator**

| $\frac{v^2-v}{2}$ | 0 | $\frac{v^2-v}{2}$ | 0 | $\frac{v^2-v}{2}$ |
|---|---|---|---|---|
| 0 | $-v$ | $-v$ | $-v$ | 0 |
| $\frac{v^2-v}{2}$ | 0 | 8 | 0 | $\frac{v^2-v}{2}$ |
| 0 | $-v$ | $-v$ | $-v$ | 0 |
| $\frac{v^2-v}{2}$ | 0 | $\frac{v^2-v}{2}$ | 0 | $\frac{v^2-v}{2}$ |

In order to extract detail information of image texture, the sum of coefficient in the fractional differential mask is not zero. In order to make the image fractional differential process has a better rotation invariant, we select four kinds of fractional differential mask operator at the same time which is in the x, y, right diagonal, left diagonal direction, we calculate the image pixels and neighborhood pixels, and then compare the four operation results, we make the maximum as pixel fractional differential gray value, finally we point-to-point stack the original image and its fractional differential diagram corresponding pixel grayscale value, at last, we obtained the gradient image of the fractional differential process as shown in Figure 12.

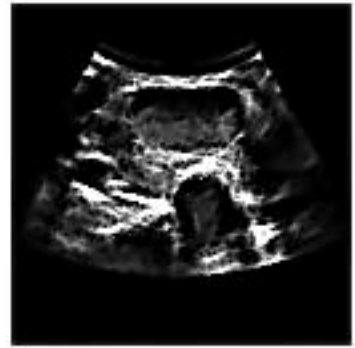

**Figure 12 Fractional gradient image**

## *Improved normalized cut of ultrasound image segmentation*

The normalized cut adopts a cut value of the image to express the objective function, through solving extreme of the objective function to realize segmentation. The features are as follows: ① It maps the problem of image segmentation to the problem of graph partitioning. ②It presents new segmentation criteria of overall situation, extract the overall effect of the images. ③ Effectively measure the dissimilarity between different groups and the overall similarity between the same groups. ④It makes the problem simplify by changing the problem into solving extensive characteristic value problem. The specific detailed introduction of the normalized cut algorithm is as follows.

If the weight-function is:

$$w_{ij} = e^{-\frac{\|F_{(i)}-F_{(j)}\|_2^2}{s_I^2}} * \begin{cases} e^{-\frac{\|F_{(i)}-F_{(j)}\|_2^2}{s_X^2}} & if \|F_{(i)}-F_{(j)}\|^2 < r \\ 0 & else \end{cases} \quad (18)$$

$F(i)$ is the gray value of pixels; $X(i)$ is the spatial coordinates of pixels; $s_I^2$ is the standard deviation of gray-scale Gaussian function; $s_X^2$ is the standard deviation of spatial distance Gaussian function; $r$ is the effective distance between two pixels, that we consider the similarity between two pixels as zero if the distance exceed. So the closer the gray values between the two pixels, the greater similarity between the two pixels and the closer the distance between the two pixels, the greater similarity is.

The normalized cut criterion not only measures the overall similarity between the different groups, but also measures the overall similarity within each group. As the principle of Ostu threshold segmentation method, the best image segmentation threshold value is the grey value which is minimum variance in the class and the biggest variance between the classes, Ncut criterion calculates the similarity between the classes, the smaller similarity between the class and the bigger similarity in the class illustrates the better segmentation results. According to the literature and considered the form of Reyleigh quotient, the rules of Ncut is transformed into solving generalized characteristic values:

$$(D - W)y = l\,Dy \qquad (19)$$

Working out Fiedler value and Fiedler vector, we finish the segmentation combined the characteristic vector of the smallest several figures. We conducted the normalized cut segmentation process after homomorphic filtering contrast enhancement, anisotropic diffusion edge-preserving smoothing and fractional differential gradient processing, the results is shown in Figure 13.

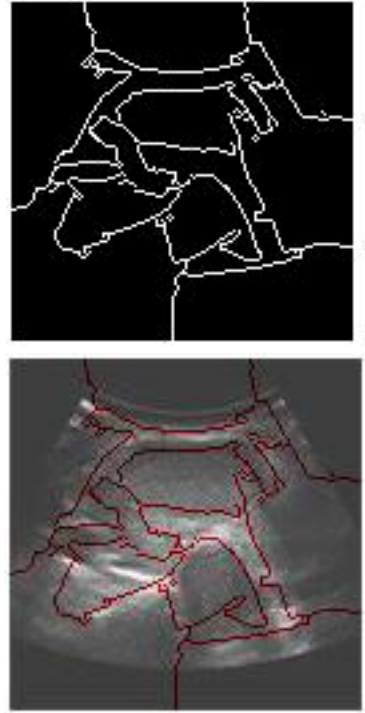

**Figure 13 Segmentation results of improved normalized cut**

Comparing the simulation result based on improved normalized cut with the six kinds of segmentation method in the previous mentioned, it is easy to find only the algorithm can produce correct segmental result on thyroid ultrasound image. Edge detecting method is not able to obtain continuous, practical value boundary (showed in Figure 8). It is very difficult to find an appropriate threshold in threshold segmentation method as a result of small gray difference between object and background (showed in Figure 9). Region splitting and merging method produces an insignificant segmental region (showed in Figure 10). Severe over-segmentation phenomenon arises in Watershed segmentation (showed in Figure 11). Practical segmental region is not able to be obtained in active contour model owing to low gray contrast (showed in Figure 12). Normalized cut based graph theory produces under-segmentation; the segmental result cannot be used for clinical diagnosis as before (showed in Figure 4). From the segmentation results (showed in Figure 7) of improved normalized cut we can see that the position of the trachea and nodule have been divided out, which provides accurate pierce position for fine needle aspiration cytology (FNAC).

# Conclusion

The biggest advantage of normalized cut is that it does not prone to small region segmentation, and it is suitable for segmentation of medical image feature. However, there is a problem to solve the normalized cut. The high memory is needed and the weight matrix calculation is large. It is easy to generate the over segmentation or less segmentation, which leads to inaccuracy in the segmentation. This paper presents an improved method of the normalized cut, introducing homomorphic filtering, anisotropic diffusion and fractional differential into the normalization process. The experimental results show that this method can extract nodules and trachea of the thyroid ultrasound images. The edge of this segmentation for fine needle aspiration cytology provides the position of piercing, assisting fine needle aspiration cytology to complete the discrimination that whether the thyroid nodule is benign or malignant.

In the procedure of using the algorithm, establishment of parameters of anisotropic diffusion model and similarity definition is the crucial to the fianl results. The parameters of anisotropic diffusion model including $\Delta t$, $n$, $a$, $b$, $c$, $l$ are selected in according to simulation experiment results. Range of iteration step $\Delta t$ is 0.06 0.3. If the step is configured smaller than 0.06,an ideal processing result cannot be obtained. If the step is configured greater than 0.3, a result image is not be true to the original. Iteration frequency $n$ is configured as 50, and $(a, b, c, l) = (0.15, 1.4, 0.015, 0.015)$. Weight matrix is used to define similarity definition and parameters of the weight matrix show parameters of similarity. Simulation experiment shows the below configuration, $Ncut = 0.065$, $\sigma_X = 0.1$, $\sigma_I = 0.3$, $r = 20$. However there still exists the problem of the algorithm's versatility and universality in the using of the algorithm, especially the problem of the setting and optimization of multiple initial parameters is also needed to deeply study.

# Competing interests

The authors including support units declare that they have no competing interests. The authors include Jie Zhao, Wei Zheng, Li Zhang, and Hua Tian. Support units include the Science Research Program of the Education Department of Hebei Province, the Open Foundation of Biomedical Multidisciplinary Research Center of Hebei University, and Main Item of Medical Science Research Plans of the Health Department of Hebei Province.

# Authors' contributions

JZ carried out the image segmentation algorithm studies, participated in the algorithm design and software programming and debugging, and drafted the manuscript. WZ carried out the image studies, participated in the experimental comparison work, and further improved the manuscript. LZ carried out the collection of early data. HT collated and studied the literature. All authors read and approved the final manuscript.

# Ethical approval and Consent

As shown in Figure 1, the name of the body is Wang gaimei and the reference number is 344641. Also, we declare that written informed consent was obtained from the patient for publication of this report and any accompanying images.

# Acknowledgements

The work is supported by Science Research Program of the Education Department of Hebei Province (2010218), Open Foundation of Biomedical Multidisciplinary Research Center of Hebei University(BM201103)and Main Item of Medical Science Research Plans of the Health Department of Hebei Province [Project No. 20120395].